%% file: SN_CycleStarGANs.tex
\documentclass[runningheads]{llncs}
\usepackage{makeidx}  
\usepackage{url}      
\usepackage{graphicx}
\usepackage{amsfonts}
\usepackage{amsmath}
\usepackage{algorithm}
\usepackage{algpseudocode}
\usepackage{subfigure}
\usepackage{multirow}
\usepackage{hyperref}
\usepackage{booktabs}
\usepackage{color}
\usepackage{textcomp}
\usepackage{caption}
\usepackage{wrapfig}
\usepackage[english]{babel}
\usepackage{blindtext}
\usepackage{romannum}
\usepackage[T1]{fontenc}
\usepackage{tabu}

\begin{document}
\frontmatter          
\titlerunning{StainGAN: Stain Style Transfer for Digital Histological Images}
\authorrunning{M.T. Shaban, C. Baur, N. Navab, S. Albarqouni}
\mainmatter              
\title{StainGAN: Stain Style Transfer for Digital Histological Images}
\author{M Tarek Shaban\inst{1}\and Christoph Baur\inst{1}\and Nassir Navab\inst{1,2}\and Shadi Albarqouni\inst{1}}
\institute{Computer Aided Medical Procedures (CAMP), Technische Universit\"at M\"unchen, Munich, Germany
\and  
Whiting School of Engineering, Johns Hopkins University, Baltimore, USA \\
\href{mailto:first.last@tum.de}{first.last@tum.de}}

\maketitle

\begin{abstract}
Digitized Histological diagnosis is in increasing demand. However, color variations due to various factors are imposing obstacles to the diagnosis process. The problem of stain color variations is a well-defined problem with many proposed solutions. Most of these solutions are highly dependent on a reference template slide. We propose a deep-learning solution inspired by CycleGANs that is trained end-to-end, eliminating the need for an expert to pick a representative reference slide. Our approach showed superior results quantitatively and qualitatively against the state of the art methods (10\% improvement visually using SSIM). We further validated our  method on a clinical use-case, namely Breast Cancer tumor classification, showing 12\% increase in AUC. Code will be made publicly available. 

\end{abstract}

\input{Intro}
\input{methodology}
\input{experiment}

\input{discussion}

\section*{Acknowledgment}
This work was partially funded by the German Research Foundation (DFG, SFB 824), and Bavarian Research Foundation (BFS, IPN2). 

\bibliography{SN_CycleStarGANs,biblio-macros}
\bibliographystyle{splncs03}
\end{document}

%% file: Intro.tex
\section{Introduction}
\label{sec:intro}

\begin{figure}	
	\subfigure[Source]{\includegraphics[width =0.23\textwidth]{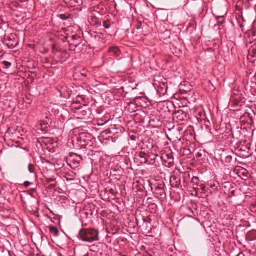}}
	\subfigure[Target]{\includegraphics[width =0.23\textwidth]{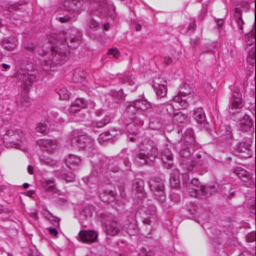} }
	\subfigure[Reinhard]{\includegraphics[width =0.23\textwidth]{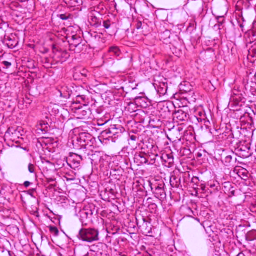}}
	\subfigure[Macenko]{\includegraphics[width =0.23\textwidth]{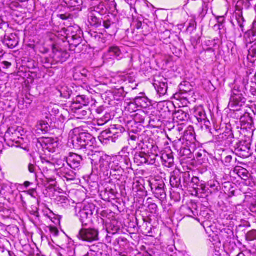}}\\
	\subfigure[Khan]{\includegraphics[width =0.23\textwidth]{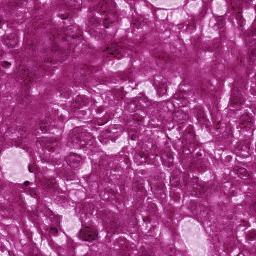}}
	\subfigure[Vahadane]{\includegraphics[width =0.23\textwidth]{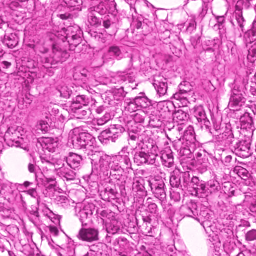}}
	\subfigure[StainGAN]{\includegraphics[width =0.23\textwidth]{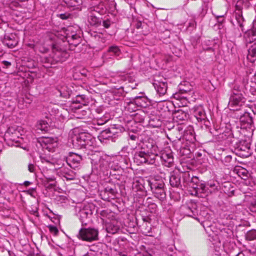}}
	\caption{Stain Normalization of various methods, the goal is to match the Target image.}
	\label{fig:visual-1}
\end{figure}

Histology relies on the study of microscopic images to diagnose disease based on the cell structures and arrangements.  Staining is a crucial part of the tissue preparation process.  The addition of staining components (mainly Hematoxylin and Eosin) transforms the naturally transparent tissue elements to become more distinguishable ( hematoxylin dyes the nuclei a dark purple color and the eosin dyes other structures a pink color). However, results from the staining process are inconsistent and prone to variability; due to differences in raw materials,  staining protocols across different pathology labs, inter-patient variabilities, and slide scanner variations as shown in \ref{fig:visual-1}. These variations not only cause inconsistencies within pathologists \cite{ismail1989observer} but it also hinders the performance of Computer-Aided Diagnosis (CAD) systems \cite{ciompi2017importance}.

Stain normalization algorithms have been introduced to overcome this well-defined problem of stain color variations. These methods can be broadly classified into three classes, \textbf{Color matching based methods} that try to match the color spectrum of the image to that of the reference template image. Reinhard et al.\cite{reinhard2001color}  align the color-channels to match that of the reference image in the lab color model, However, this can lead to improper color mapping, as the same transformation is applied across the images and does not take into account the independent contribution of stain dyes to the final color.

Further, there are \textbf{Stain-separation methods} where normalization is done on each staining channel independently. For instance,  Macenko et al. \cite{macenko2009method} find the stain vectors by transforming the RGB to the Optical Density (OD) space. On the other hand, the method proposed by Khan et al. \cite{khan2014nonlinear}  estimate the stain matrix based on a color-based classifier that assigns every pixel to the appropriate stain component. According to Babak et al \cite{bejnordi2016stain}, these methods do not take the spatial features of the tissue structure into account, which leads to improper staining. Nevertheless, most of the methods in this class rely on an expertly picked reference template image, which has a major effect on the outcome of the methods as we show later. The third class of methods is the \textbf{Pure learning based approaches}, that handle the problem of stain normalization as a style-transfer problem, BenTaieb et al.\cite{bentaieb2017adversarial} handle the problem using auxiliary classifier GAN with an auxiliary task on top (classifier or segmentation). Our method (StainGAN)  is also GAN based but as opposed to BenTaieb does not require to be trained for a specific task in order to achieve stain style transfer.

We propose a method based on Generative adversarial networks (GANs) that not only eliminates the need for the reference image but also achieves high visual similarity to the target domain, making it easier to get rid of the stain variations thus improving the diagnosis process for both the pathologist and CAD systems. StainGAN is based on the Unpaired Image-to-Image Translation using Cycle-Consistent Adversarial Networks (CycleGAN)\cite{zhu2017unpaired}.  Cycle-consistency allows the images to be mapped to different color-model but preserves the same tissue structure. We evaluate our approach against the state of the art methods quantitatively and indirectly through a breast tumor classification. Our contributions include 1) Using Style-transfer for the classic stain normalization problem and achieving better results quantitatively than the state of the art methods. 2) Removing the need for a manually picked reference template, as our model learn the whole distribution. 3) Providing a benchmark for comprehensive comparison between the proposed StainGAN method and most of the state of the art methods.

%% file: methodology.tex
\begin{figure}[t]
	\centering
	{\includegraphics[width =0.7\textwidth]{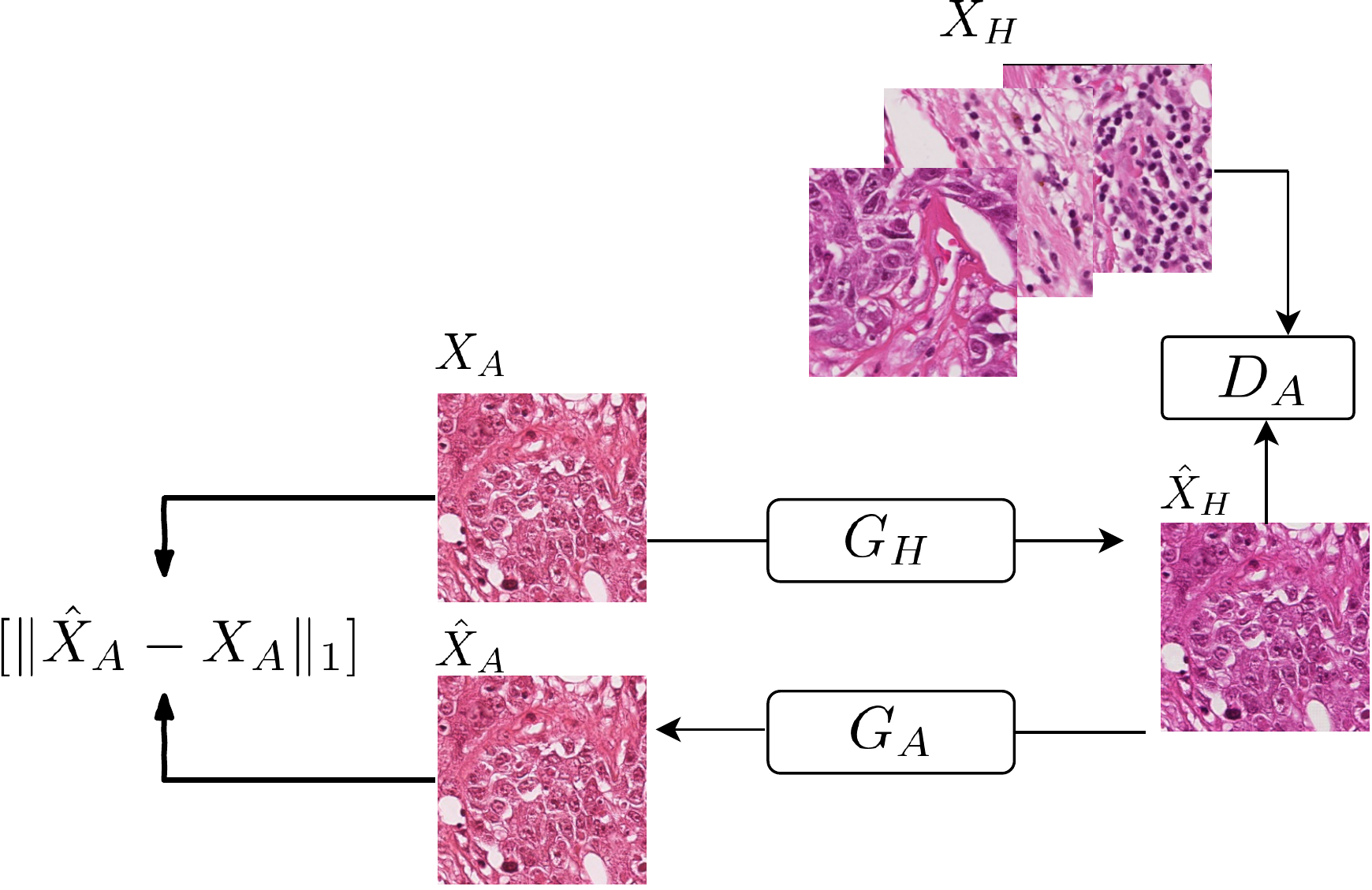}}
	\caption{Our StainGAN framework: Images are mapped to domain H and then back to domain A to ensure structure perseverance. Same process is done in reverse direction from domain H to domain A. }
	\label{fig:framework}
\end{figure}

\section{Methodology}
\label{sec:method}
Our framework, as depicted in Fig.~\ref{fig:framework}, employs the CycleGAN concept~\cite{zhu2017unpaired} to transfer the H\&E Stain Appearance between different scanners,~\emph{i.e} from Hamamatsu (H) to Aperio (A) Scanner, without the need of paired data from both domains. The model consists of two generator and discriminator pairs, the first pair ($G_H$ and $D_H$), tries to map images from domain $A$ to domain $H$ $G_H: \mathcal{X}_{A}\rightarrow \mathcal{X}_{H}$. While the Generator $G_H$ tries to generate images that match domain $H$, the discriminator $D_H$ tries to verify if images come from the real domain $H$ or the fake generated ones. The other pair ($G_A$ and $D_A$), undergoes the same process in the reverse direction, $G_A: \mathcal{X}_{H}\rightarrow \mathcal{X}_{A}$, as
\begin{equation}
\label{eq:1}
\hat{X}_H = G_H(X_A; \theta_H), \hspace{5pt} \hat{X}_A = G_A(\hat{X}_H; \theta_A), \hspace{5pt} \text{s.t.} \hspace{5pt} d(X_A, \hat{X}_A) \leq \epsilon,
\end{equation}
and 
\begin{equation}
\label{eq:2}
\hat{X}_A = G_A(X_H; \theta_A), \hspace{5pt} \hat{X}_H = G_H(\hat{X}_A; \theta_H), \hspace{5pt} \text{s.t.} \hspace{5pt}  d(X_H, \hat{X}_H) \leq \epsilon,
\end{equation}
where $d(\cdot, \cdot)$ is a distance metric between the given image and the reconstructed one, so-called a \textit{Cycle-Consistency} constraint, and both $\theta_A$ and $\theta_H$ are the model parameters.

To achieve this, the models are trained to meet the following objective function,
\begin{equation}
\label{eq:full}
\mathcal{L} = \mathcal{L}_{Adv} + \lambda \mathcal{L}_{Cycle},
\end{equation}
where $\mathcal{L}_{Adv}$ is the adversarial loss, $\mathcal{L}_{Cycle}$ is the cycle-consistency loss, and $\lambda$ is a regularization parameter. 

The adversarial loss tries to match the distribution of the generated images to that of the target domain (\textit{Forward Cycle}), and match the distribution of the generated target domain back to the source domain (\textit{Backward Cycle}) as 
$\mathcal{L}_{Adv} = \mathcal{L}_{GAN}^{A} + \mathcal{L}_{GAN}^{H},$
where $\mathcal{L}_{GAN}^{A}$ is given as 
\begin{equation}
\label{eq:gengan}
\begin{split}
 \mathcal{L}_{GAN}^{A}(G_A,D_A,X_H,X_A) & = \mathbb{E}_{X_A\sim p_{data}(X_A)}[\log D_A(X_A)]\\
& + \mathbb{E}_{X_H\sim p_{data}(X_H)}[\log (1-D_A(G_A(X_H; \theta_A))]. 
\end{split}
\end{equation}

The Cycle-consistency loss ensures that generated images preserve similar structure as in the source domain. This loss goes in both directions forward and backward cycles to assure stability. 
\begin{equation}
\label{eq:cycle}
\begin{split}
\mathcal{L}_{Cycle}(G_A,G_H,X_A,X_H) & = \mathbb{E}_{X_A\sim p_{data}(X_A)}[\|G_A(G_H(X_A;\theta_H);\theta_A)) - X_A\|_1]\\
& + \mathbb{E}_{X_H\sim p_{data}(X_H)}[\|G_H(G_A(X_H;\theta_A);\theta_H)) - X_H\|_1],
\end{split}
\end{equation}
where $\|\cdot\|_1$ is the $\ell_1$-norm.
\paragraph{Network Architectures.} In our model, ResNet~\cite{he2016deep} and $70\times70$ PatchGAN~\cite{isola2017image} are employed as network architecture for the generators and the discriminators, respectively. 

%% file: experiment.tex
\section{Experiments and Results}
To have a fair comprehensive  comparison, we evaluate our model against the state of the art methods of Reinhard\cite{reinhard2001color} Macenko\cite{macenko2009method}, Khan\cite{khan2014nonlinear}, and Vahadane\cite{vahadane2016structure} as follows; i) \textit{Quantitatve comparison} between the visual appearances of stained images, and the effect of varying reference slides, ii) \textit{Quantitative comparison} of the Hematoxylin and Eosin stain vectors separation, and iii) \textit{Use-case experiment}, stain normalization as a preprocessing step in the pipeline of tumor classifier.

\subsection{Stain Transfer}
The goal is to be able to map the patches from scanner $A$(Aperio) to that of Scanner $H$(Hamamatsu), the dataset includes same tissue sections scanned with both scanners, Slides from scanner A are normalized to match scanner H, then compared with the real slides of Scanner H (ground truth). Results are evaluated using various similarity matrices. Additionally, Stain Vectors are extracted using Ruifrok's method \cite{ruifrok2001quantification} and compared to that of the ground truth.
Visual results of the of the various methods are illustrated in \ref{fig:visual-1}. It is clear that our results are more close visually to the target images.

\paragraph{\textbf{Dataset.}}
The dataset is publicly available as part of the MITOS-ATYPIA 14 challenge \footnote{https://mitos-atypia-14.grand-challenge.org}. Dataset consists of $284$ frames at $X20$ magnification which are stained with standard Hematoxylin and Eosin (H\&E) dyes. Same tissue section has been scanned by two slide scanners: Aperio Scanscope XT and Hamamatsu Nanozoomer 2.0-HT. Slides from both scanners resized to have equal dimensions of ($1539\times1376$). Rigid registration was performed to eliminate any misalignment. For the training set, we extract 10,000 random patches from the first 184 full slide of both scanners. For evaluation, $500$ same view section patches of $256\times256$ were generated from each of the two scanners from the last 100 full slides (unseen in the training set), patches from the Scanner H are used as the ground truth.

\paragraph{\textbf{Implementation.}} 
For the state of the art methods, a refrence slide was picked carefully. Our method does not require a reference slide as it learns the distribution of the whole data, model was trained using $10,000$ random unpaired patches from both scanners for 26 epochs with the regularization parameter set to $\lambda$= 10, learning rate is set to $0.0002$, Adam optimizer with a batch size of $4$ is used. Hardware of GeForce GTX TITAN X 12GB and Pytorch framework were used.

\paragraph{\textbf{Evaluation Metrics.}} Results are compared to the ground truth using four similarity measures: Structural Similarity index (SSIM)\cite{wang2004image}, Feature Similarity Index for Image Quality Assessment (FSIM) \cite{zhang2011fsim}, Peak Signal-to-Noise Ratio (PSNR) and Pearson correlation coefficient similarity\cite{ahlgren2003requirements}.
On the other hand, Euclidian norm distance as reported in~\cite{bejnordi2016stain} was used to evaluate stain vectors against the ground truth.

\paragraph{\textbf{Results.}} As reported in Table \ref{tab:similarity}, our results significantly outperform the state-of-the-art methods in all similarity metrics ($p < 0.01$). Further, it has shown a better stain seperability compared to the ground-truth stain vectors as shown in Table. \ref{tab:stainvector}. 

\begin{table}[t]
	\caption{Stain Transfer Comparison: Mean $\pm$ Standard Deviation}
	\label{tab:staintransfer}
	\centering
	\begin{tabular*}{\textwidth}{@{}l| *{4}{@{\extracolsep{\fill}}c}} 
		
		\toprule
		
		Methods & SSIM & FSSIM  & Pearson Correlation & PSNR  \\
		\hline
		Reinhard~\cite{reinhard2001color} & $0.58  \pm 0.10$ & $0.67  \pm 0.05$  & $0.51  \pm 0.21$   & $13.4  \pm 1.61$  \\
		Macenko~\cite{macenko2009method} & $0.56  \pm 0.12$ & $0.67  \pm 0.05$  & $0.45  \pm 0.22$ & $14.0  \pm 1.68$   \\	
		Khan~\cite{khan2014nonlinear} & $0.67 \pm 0.11$   & $0.71 \pm 0.05$  & $0.54  \pm 0.20$  &$16.3  \pm 2.11$ \\
		Vahadane~\cite{vahadane2016structure} & $0.65 \pm 0.13$ & $0.71  \pm0.06$  & $0.53  \pm 0.22$ & $14.2  \pm 2.13$ \\
		StainGAN &  \textbf{0.71 $\pm$ 0.11} &  \textbf{0.73 $\pm$ 0.06}   & \textbf{0.56  $\pm$ 0.22}  & \textbf{17.1  $\pm$ 2.50}  \\			
		
		\bottomrule
	\end{tabular*}
	\label{tab:similarity}
\end{table}

\begin{table}[t]
	\caption{Stain Vectors Comparison: Mean $\pm$ Standard Deviation (the lower the better) and Processing time taken to normalize the 500 images}
	\centering
	\begin{tabular*}{0.68\textwidth}{@{}l|*{2}{@{}c}|@{}c}
		
		\toprule
		Methods &  \multicolumn{2}{c}{Staining Separation} \vline & Processing time \\
		\midrule	
		 & $S_{H}$ & $S_{E}$ & Time (sec) \\
		
		Reinhard~\cite{reinhard2001color}  & $20.7 \pm 4.85$  &  $18.0 \pm 3.79$  &  \textbf{9.80} \\
		Macenko~\cite{macenko2009method}  & $21.05\pm 4.11$ &   $12.2 \pm 3.14$ &  63.63 \\	
		Khan~\cite{khan2014nonlinear}  & $21.2 \pm 3.65$   &  $12.0 \pm 2.73$  & $2196.25$\\
		Vahadane~\cite{vahadane2016structure} & $20.0 \pm 4.57$ &  $12.9 \pm 3.45$ & $582.94$   \\
		StainGAN &  \textbf{18.0$ \pm$ 4.84} &  \textbf{ 10.0$ \pm$ 3.32} & $74.55$ \\				
		\bottomrule
	\end{tabular*}
	\label{tab:stainvector}
\end{table}

\paragraph{\textbf{Slide Reference Sensitivity.}} 
We ran the same experiment as in the previous one, but with three different reference images shown in changes of the SSIM score with respect to the reference images, results are reported in Fig.\ref{fig:ROC}, which shows that the conventional methods are sensitive to the reference image.

\subsection{Use-Case Application}

Stain-normalization is a mandatory preprocessing step in most CAD systems and has proven to  increase performance\cite{ciompi2017importance}. The aim of this experiment is to show the performance of stain normalization in the context of breast cancer classification, model is trained with patches from the lab 1,  testing is done with patches from lab 2 (different staining appearances). We normalize the test-set to match lab 1 and compare results accordingly.

\paragraph{\textbf{Dataset.}}
Camelyon16 challenge \footnote{https://camelyon16.grand-challenge.org/} consisting of 400 whole-slide images collected in two different labs in Radboud University Medical Center (lab 1) and University Medical Center Utrecht (lab 2). Otsu thresholding was used to remove the background, Afterwards, $40,000$ $256\times256$ patches were generated on the x40 magnification level, $30,000$ were used for training and $10,000$ used for validation from lab 1 and $10,000$ patches were generated for testing from lab 2. 

\paragraph{\textbf{Implementation.}}
The classifier network is composed of three convolutional layers with a ReLU activation, followed by a max-pooling layer turning the given image into a logit vector. Classifier has been trained for $30$ epochs with RMSprop optimizer and binary cross-entropy loss.
For the stain normalization, representative reference slide form lab 2 was picked for the conventional methods, Our method was trained using the same parameters from the previous experiment on training set: $68000$ patches from both labs.

\paragraph{\textbf{Evaluation Metrics.}} 
To evaluate the classifier performance, we report the Area Under the Curve (AUC) of the Receiver Operating Characteristics (ROC).

\begin{figure}[t]
	\centering 
	\subfigure[]{\includegraphics[width =0.42\textwidth]{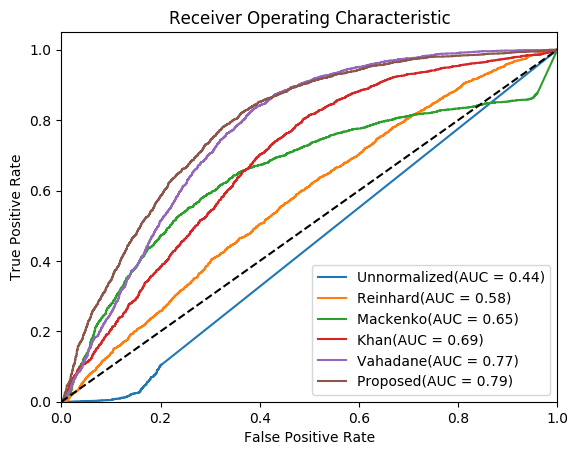}}
	\subfigure[]{\includegraphics[width =0.55\textwidth]{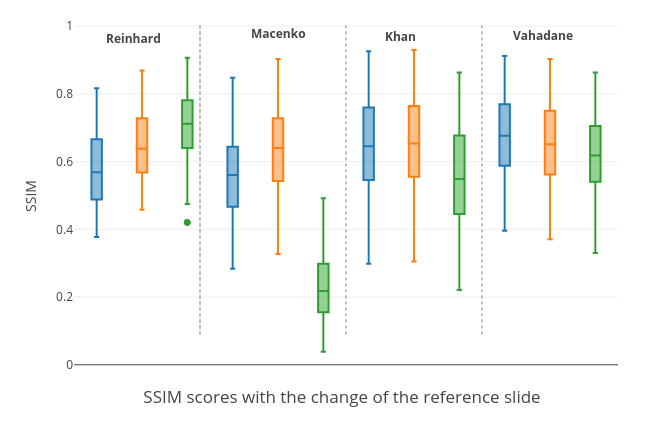}}
	\caption{(a) ROC curves of the test-set pre-processed using different stain normalization methods. (b) Box plot shows the variation of the SSIM metric due to the improper selection of reference slide.}
	\label{fig:ROC}
\end{figure}

\paragraph{\textbf{Results.}} 
The ROC curves and AUC of different classifiers, trained on different stain normalization methods, are presented in Fig.\ref{fig:ROC}. Our proposed StainGAN method shows a relative improvement of ($80\%$, $36\%$, $22\%$, $15\%$, and $3\%$) on AUC over the Un-normalized, Reinhard, Macenko, Khan, and Vahadane, respectively.

%% file: discussion.tex
\section{Discussion and Conclusion}

In the paper, we presented StainGAN as a novel method for the stain normalization task. Our experiments revealed that our method significantly outperforms the state of the art. The visual appearance of different methods can be seen in Fig.\ref{fig:visual-1}. It clearly shows that images normalized with StainGAN are very similar to the ground truth. Further, our StainGAN method has been validated in a clinical use-case, namely Tumor Classification, as a pre-processing step showing a superior performance.  
Moreover, the processing time of our method is on par with Macenko as reported in Table. \ref{tab:stainvector}. We believe that end-to-end learning based approaches are ought to overtake the classic stain normalization methods. Yet, there is still more room for improvement, for example, we can think of Unified representation, similar to \cite{choi2017stargan}, that can map many to many stain style domains. 